\documentclass[letterpaper, 10 pt, conference]{ieeeconf}  

\IEEEoverridecommandlockouts                              

\usepackage{cite}
\usepackage{graphics} 
\usepackage{epsfig} 
\usepackage{amsmath} 
\usepackage{amssymb}  
\usepackage{pifont}
\usepackage{booktabs,tabularx}
\usepackage{url}
\usepackage{hyperref}
\usepackage{xcolor}
\newcommand{\greencheck}{{\color{teal}\ding{51}}}

\newcommand{\xmark}{\color{purple}\ding{55}}

\title{\LARGE \bf
Mind and Motion Aligned: A Joint Evaluation IsaacSim Benchmark for Task Planning and Low-Level Policies in Mobile Manipulation
}

\author{Nikita Kachaev$^{1}$, Andrei Spiridonov$^{1}$, Andrey Gorodetsky$^{1}$, Kirill Muravyev$^{4,2}$, Nikita Oskolkov$^{2}$, Aditya Narendra$^{2}$,\\Vlad Shakhuro$^{1,3}$, Dmitry Makarov$^{2,4}$, Aleksandr I. Panov$^{1,2}$, Polina Fedotova$^{5,6}$ and Alexey K. Kovalev$^{1,2}$
\thanks{$^{1}$AIRI, Moscow, Russia $^{2}$Moscow Institute of Physics and Technology (MIPT), Dolgoprudny, Russia $^{3}$Lomonosov Moscow State University, Moscow, Russia $^{4}$Federal Research Center ``Computer Science and Control'' of Russian Academy of Sciences (FRC CSC RAS), Moscow, Russia  $^{5}$Sberbank, Robotics Center, Moscow, Russia $^{5}$Skoltech, Moscow, Russia.}%
}

\begin{document}

\maketitle
\thispagestyle{empty}
\pagestyle{empty}

\begin{abstract}
Benchmarks are crucial for evaluating progress in robotics and embodied AI. However, a significant gap exists between benchmarks designed for high-level language instruction following, which often assume perfect low-level execution, and those for low-level robot control, which rely on simple, one-step commands. This disconnect prevents a comprehensive evaluation of integrated systems where both task planning and physical execution are critical. To address this, we propose \textbf{Kitchen-R}, a novel benchmark that unifies the evaluation of task planning and low-level control within a simulated kitchen environment. Built as a digital twin using the Isaac Sim simulator and featuring more than 500 complex language instructions, Kitchen-R supports a mobile manipulator robot. We provide baseline methods for our benchmark, including a task-planning strategy based on a vision-language model and a low-level control policy based on diffusion policy.
We also provide a trajectory collection system. Our benchmark offers a flexible framework for three evaluation modes: independent assessment of the planning module, independent assessment of the control policy, and, crucially, an integrated evaluation of the whole system. Kitchen-R bridges a key gap in embodied AI research, enabling more holistic and realistic benchmarking of language-guided robotic agents.
\end{abstract}

\section{INTRODUCTION}
Benchmarks are widely used in natural language processing~\cite{wang-etal-2018-glue} and computer vision~\cite{krishna2017visual} to assess the progress of models and compare their performance. In robotics, benchmarks based on simulators are also common~\cite{gong2023arnold,pmlr-v205-li23a,Pumacay-RSS-24}, often serving the dual purpose of evaluating and training models. A key consideration in this context is the accurate simulation of low-level actions, which facilitates the transfer of results to real-world robots.

In recent years, large language models (LLMs) and vision-language models (VLMs) have been widely adopted for task planning and instruction following in robotics~\cite{ahn2022can,kovalev2022application,sarkisyan2023evaluation,10160591,10161317,yang2024guiding,onishchenkolookplangraph,grigorev2025verifyllm,pchelintsev2025lera}, a direction closely tied to embodied AI. Evaluating such approaches requires specialized benchmarks that account for the challenges of deriving task plans from language instructions~\cite{puig2018virtualhome,shridhar2020alfred,padmakumar2022teach,choi2024lotabench,li2024embodied}. However, these benchmarks often overlook the actual execution of tasks in real-world environments.

To fully evaluate an embodied AI system, it is essential to assess both the quality of task planning and the quality of task execution under conditions that are as realistic as possible. Modern benchmarks for instruction following primarily focus on evaluating task planning, often assuming that atomic task execution is always successful~\cite{puig2018virtualhome} or simplifying it, e.g. by reducing interactions to the target object mask manipulation~\cite{shridhar2020alfred,padmakumar2022teach}. Meanwhile, popular robotics benchmarks emphasize realistic action execution but rely on short, one-step language instructions~\cite{gong2023arnold} or omit language-based guidance entirely~\cite{pmlr-v205-li23a}.

\begin{figure}[t]
    \centering
    \includegraphics[width=\linewidth]{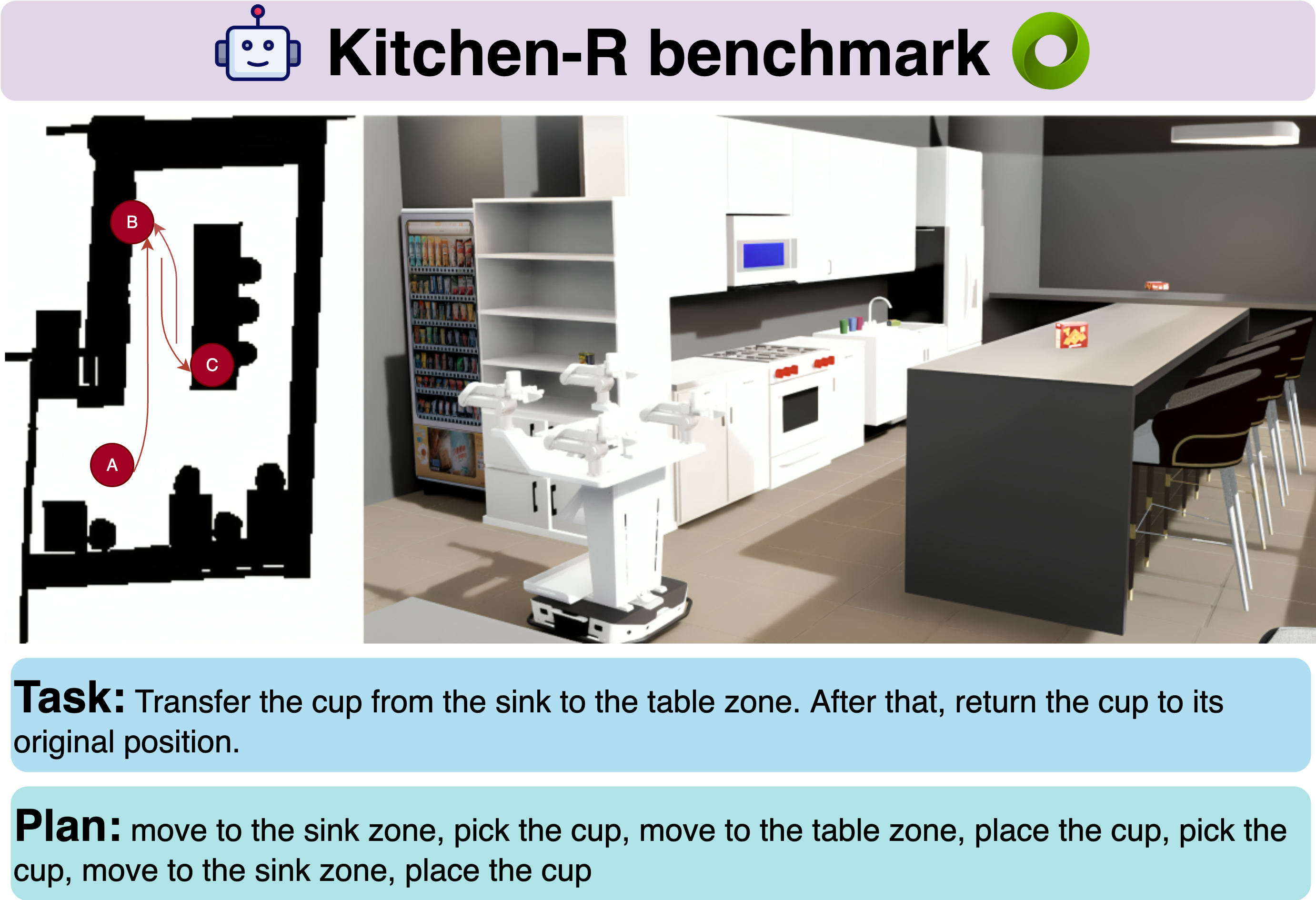}
    \caption{Overview of a digital twin of a kitchen in the Kitchen-R benchmark, including its occupancy map with an annotated motion trajectory, a language instruction, and the corresponding high-level plan generated by a VLM.}
    \label{fig:kithenr}
\end{figure}

In this paper, we address this gap by proposing a unified approach to instruction following and robotics benchmarks. We introduce \textbf{Kitchen-R} (where ``R'' denotes robots), a benchmark where a mobile robot (mobile ALOHA~\cite{fu2024mobile}) executes language instructions in a simulated digital twin of a real kitchen (see Figure~\ref{fig:kithenr}). Built on the Isaac Sim simulator\cite{nvidia_isaac_sim_500_2025} for realistic action execution and augmented with more than 500 language instructions, our benchmark includes: a VLM-based planning baseline for language instruction following, and a diffusion policy~\cite{chi2023diffusion} baseline for low-level mobile robot control. Additionally, we provide a robot trajectory collection system. Kitchen-R supports three evaluation modes: 1) independent evaluation of the task planning module; 2) independent evaluation of the low-level control policy; 3) integrated evaluation of both task planning and control modules. The Kitchen-R benchmark was successfully used for data collection and validation in the Embodied AI track of the AIJ Contest 2024\footnote{\url{https://dsworks.ru/group/ai-journey-contest-2024}}. The benchmark provides a unified testbed and baselines to facilitate the development of robust agents capable of executing complex language commands in realistic environments.

To summarize, our key contributions are:
\begin{enumerate}
    \item \textbf{Kitchen-R benchmark:} a digital twin kitchen environment with more than 500 language instructions for Embodied AI research;
    \item \textbf{Baseline methods:} VLM-based task planning baseline and diffusion policy for low-level robot control;
    \item \textbf{Flexible data collection and evaluation framework} supporting modular assessment of system components.
\end{enumerate}

\section{RELATED WORK}
At present, there are few realistic simulator–based frameworks that execute high-level action plans with feedback in the loop. Arnold~\cite{gong2023arnold} targets language-grounded manipulation with continuous object states and carefully designed tasks probing generalization across goals, scenes, and objects; however, it ships a fixed task set and does not prioritize extensible large-scale trajectory collection. Grutopia~\cite{kachaevwang2024grutopiadreamgeneralrobots} pursues city-scale breadth with diverse capabilities, but its emphasis is on broad social and navigation scenarios rather than tightly integrated plan–execution loops for realistic household manipulation. OmniGibson, introduced via the BEHAVIOR-1K suite~\cite{kachaevli2024behavior1khumancenteredembodiedai}, provides realistic scenes and rich object states, yet does not natively offer the trajectory collection pipelines needed for behavior cloning and systematic feedback-aware evaluation. 

Benchmarks such as VirtualHome~\cite{puig2018virtualhome}, ALFRED~\cite{shridhar2020alfred}, TEACh~\cite{padmakumar2022teach}, EAI~\cite{li2024eai} and LoTa-Bench~\cite{choi2024lotabench} primarily assess deriving task plans from language under simplified physics or with assumed reliable primitives. While these works advance plan understanding and long-horizon composition, they provide limited visibility into how planning errors interact with controller robustness in photorealistic simulation. The COLOSSEUM~\cite{pumacay2024colosseum} stresses robustness under environmental perturbations across diverse manipulation tasks, and BEHAVIOR-1K~\cite{kachaevli2024behavior1khumancenteredembodiedai} defines a large suite of everyday activities and assets enabling long-horizon scenarios. Yet language-conditioned planning with feedback-in-the-loop execution is not the central emphasis in these benchmarks.

The Kitchen-R  benchmark situates itself at the intersection of these strands by enabling independent evaluation of planner and policy, as well as integrated end-to-end evaluation, with built-in trajectory logging and photorealistic execution in Isaac Sim (see Table \ref{tab:benchmark_comparison_full}).

\begin{table}[t]
\centering
\caption{Comparison of various frameworks and benchmarks.}
\resizebox{\linewidth}{!}{
\begin{tabular}{lccccccccccc}
\toprule
\textbf{Benchmark} & \rotatebox{90}{\textbf{Lang. planning}} & \rotatebox{90}{\textbf{Multiple Cameras}} & \rotatebox{90}{\textbf{LiDAR}} & \rotatebox{90}{\textbf{Data Collection}} & \rotatebox{90}{\textbf{Add Scenes}} & \rotatebox{90}{\textbf{Add Objects}} & \rotatebox{90}{\textbf{Add Robots}} & \rotatebox{90}{\textbf{ROS Support}} & \rotatebox{90}{\textbf{Navigation}} & \rotatebox{90}{\textbf{Manipulation}} \\
\midrule
OmniGibson~\cite{kachaevli2024behavior1khumancenteredembodiedai} & \xmark & \greencheck & \greencheck & \xmark & \xmark & \xmark & \greencheck & \xmark & \greencheck & \greencheck \\
GRUtopia~\cite{kachaevwang2024grutopiadreamgeneralrobots} & \greencheck & \greencheck & \xmark & \greencheck & \xmark & \greencheck & \xmark & \xmark & \greencheck & \greencheck \\
Arnold~\cite{gong2023arnold} & \greencheck & \xmark & \xmark & \xmark & \xmark & \greencheck & \xmark & \xmark & \xmark & \greencheck \\
COLOSSEUM~\cite{pumacay2024colosseum} & \xmark & \xmark & \xmark & \greencheck & \xmark & \xmark & \xmark & \xmark & \xmark & \greencheck \\
VirtualHome~\cite{puig2018virtualhome} & \greencheck & \greencheck & \xmark & \xmark & \xmark & \xmark & \xmark & \xmark & \greencheck & \xmark \\
ALFRED~\cite{shridhar2020alfred} & \greencheck & \greencheck & \xmark & \xmark & \xmark & \xmark & \xmark & \xmark & \greencheck & \xmark \\
TEACh~\cite{padmakumar2022teach} & \greencheck & \greencheck & \xmark & \xmark & \xmark & \xmark & \xmark & \xmark & \greencheck & \greencheck \\
EAI~\cite{li2024eai} & \greencheck & \xmark & \xmark & \xmark & \xmark & \xmark & \xmark & \xmark & \greencheck & \xmark \\
LoTa\mbox{-}Bench~\cite{choi2024lotabench} & \greencheck & \xmark & \xmark & \xmark & \xmark & \xmark & \xmark & \xmark & \greencheck & \xmark \\
\textbf{Kitchen\mbox{-}R (our)} & \greencheck & \greencheck & \greencheck & \greencheck & \greencheck & \greencheck & \greencheck & \greencheck  & \greencheck & \greencheck \\
\bottomrule
\end{tabular}
}
\label{tab:benchmark_comparison_full}
\end{table}

\section{PROBLEM FORMULATION}
We break down the general problem of following instructions with a mobile manipulator into two subproblems: task planning and mobile manipulation. The task planning subproblem requires a high-level control system based on a VLM. This system must: 1) recognize and understand complex English-language text instructions related to routine tasks; 2) generate task plans for following these instructions in a virtual environment, considering its specific features and limitations. The system's input consists of a natural-language instruction for the robot to perform navigation and object manipulation tasks, along with a top-down view of the scene. Its output is the corresponding task plan.

The mobile manipulation subproblem requires a low-level control system. This system must generate a trajectory for the mobile manipulator where each point consists of 10 values: $(v_{cmd}, \omega_{cmd}, x, y, z, q_w, q_x, q_y, q_z, g)$. Here, $v_{cmd}$ and $\omega_{cmd}$ are the mobile manipulator base's linear and angular velocities; $x$, $y$, and $z$ are the end-effector's offset coordinates; $q_w$, $q_x$, $q_y$, and $q_z$ represent the end-effector's rotation as a quaternion; and $g$ is the gripper opening (from $0$ for closed to $1$ for open). The system's input is a decomposed natural-language task instructing the mobile manipulator to perform navigation and object manipulation, along with visual information from two on-board cameras: one mounted on the gripper and a central camera on the base. The system's output is a movement trajectory that predicts the mobile manipulator's actions for several steps ahead (16 in our setting).

\begin{figure*}[t]
    \centering
    \includegraphics[width=1\linewidth]{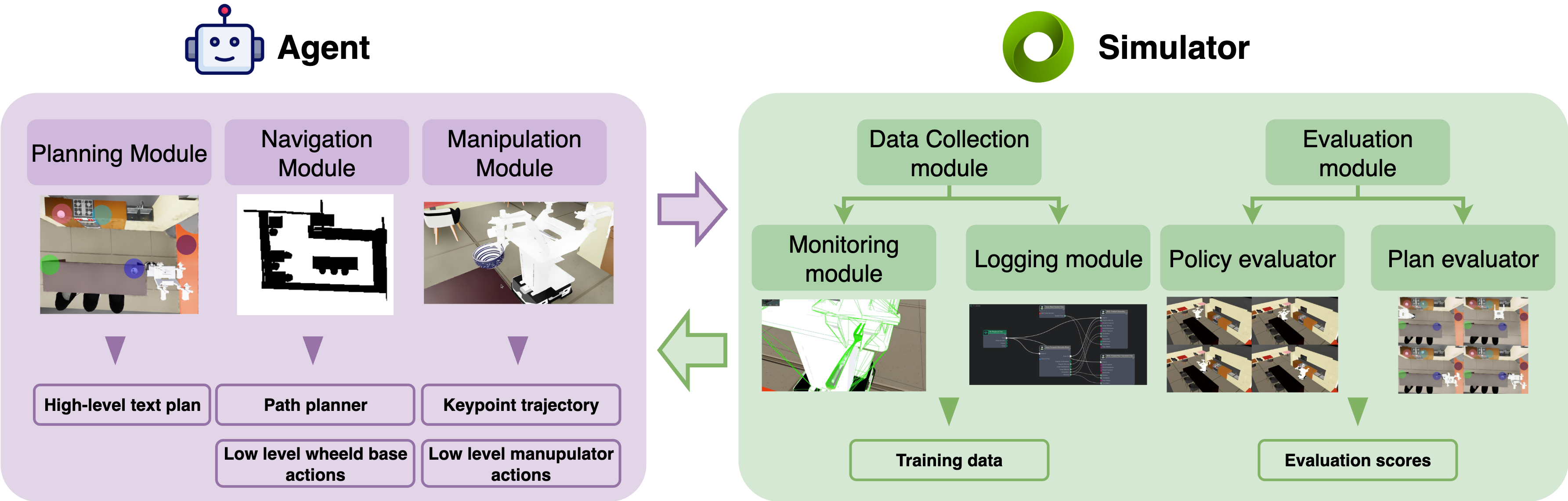}
    \caption{Diagram of the Kitchen-R benchmark. It illustrates the core modules for data collection and curation used to train the agent, the agent architecture enabling end-to-end planing, manipulation and navigation, and the modules for its evaluation in the simulator.}
    \label{fig:framework_scheme}
\end{figure*}

\section{METRICS}
For the offline validation of the task planning subproblem, we use the Exact Match (EM) metric. EM is calculated as the average of the per-plan accuracy:
\begin{equation}
    \text{EM}=\frac{1}{N}\sum_{i=1}^N\frac{PS_i}{N_i},
\end{equation}
where $N$ is the total number of plans, $\text{PS}_i$ is the number of correctly predicted steps in the plan $i$, and $N_i$ is the total number of steps in the ground-truth plan $i$.
Since a plan step is a text string (a sequence of characters), it is considered correctly predicted only if a character-by-character comparison with the correct plan step shows an exact match. A high EM score indicates that the model performs well on the task of predicting the robot's task plan.

For the offline validation of the mobile manipulation subproblem, we use the Mean Squared Error (MSE) between the predicted trajectory and the expert trajectory. The overall performance is calculated by averaging the MSE across all predicted trajectories.

For the overall system evaluation, we define a composite metric, $P$, that combines the performance of both the task planning and mobile manipulation subproblems:
\begin{equation}
    \text{P}=\frac{1}{\text{EM}}+\frac{1}{N}\sum_{i=1}^{N}(\frac{1}{n_i}\sum_{j=1}^{n_i}\text{MSE}_j),
\end{equation}
where $N$ is the total number of instructions, $n_i$ is the number of tasks for instruction $i$.

For the online joint validation of the task planning and mobile manipulation subproblems in the simulator, we use the EM and Success Rate (SR) metrics. The SR is equal to 1 for a decomposed task if the task is completed successfully within the 120-second time limit for model inference. Success is defined as follows. For a navigation task: the robot's base geometric center is within 10 cm of the target position. For a manipulation task: the geometric center of the manipulated object is within 5 cm of its target position.

The overall process of online validation follows three steps: 1) the task planning model takes a high-level instruction as input and decomposes it into a sequence of tasks. This predicted plan is compared step-by-step to the ground-truth plan to calculate the EM metric; 2) to isolate planning errors from execution errors, the ground-truth plan is always used for execution. This allows the mobile manipulation subsystem to be evaluated fairly, even if the task planner fails; 3) the mobile manipulation system executes each task in the plan. It has a maximum of 120 seconds to complete each task. The simulator evaluates the SR criteria after every action. If the success criteria are met, the task is marked as successful ($\text{SR} = 1$) and the completion time is recorded. The system then proceeds to the next task in the sequence.
The final performance metric is calculated as:
\begin{equation}
   \text{M}=\frac{1}{N}\sum_{i=1}^{N}(\frac{1}{n_i}\sum_{j=1}^{n_i}(\text{EM}_j + \text{SR}_j), 
\end{equation}
where $N$ is the total number of instructions, $n_i$ is the number of tasks for instruction $i$.

\section{KITCHEN-R BENCHMARK}
In this paper, we propose Kitchen-R, a benchmark for the joint evaluation of task planning and low-level mobile manipulation policies. The overall structure of the benchmark is presented in Figure~\ref{fig:framework_scheme}.
To streamline dataset creation, we propose a modular framework built on the physically and photorealistic simulator Nvidia Isaac Sim~\cite{nvidia_isaac_sim_500_2025} that enables collecting trajectories for training policies via behavior cloning methods and executing high-level plans using ground-truth controllers or trained low-level policies. The framework uses a modular architecture: each module takes a specific input data type and performs a concrete task while interacting with other modules. This makes it possible to replace and improve individual components with minimal impact on the rest.

\subsection{Policy evaluator}
\begin{figure}[t]
    \centering
    \includegraphics[width=1.0\linewidth]{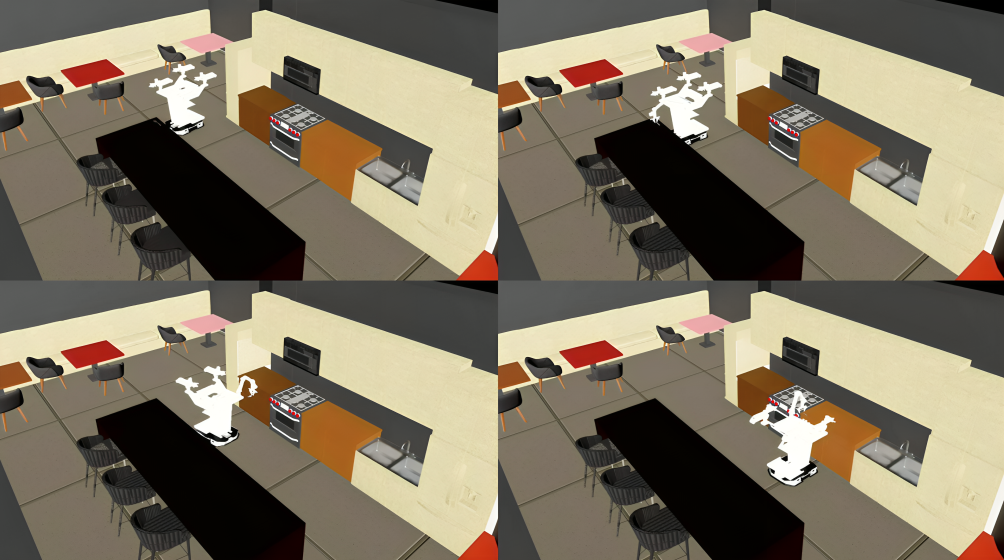}
    \caption{An example of a robot performing the task ``\texttt{Move the fork to the green zone}``. To reduce the scene’s visual complexity for data collection, textures and rendering are simplified.}
    \label{fig:example_collection}
    \vspace{-20pt}
\end{figure}

A policy evaluator is used to execute the high-level plan. It takes as input a plan decomposed into sub-tasks from the high-level task planner. The evaluator then retrieves the required data from the simulator and sequentially invokes the navigation and manipulation modules to carry out the plan. Figure \ref{fig:example_collection}  demonstrates an example of executing a high-level plan decomposed into sub-tasks in the simulator. 

\subsection{Navigation Module}
For each subtask, the policy evaluator determines the target coordinates and passes them to the path planner together with the robot’s current pose. The policy evaluator also reads from the file a precomputed scene map in the form of a 2D occupancy grid and passes it to the planner. The occupancy map is recomputed every $N$ steps or specified once initially and left unchanged.

The navigation module consists of a path planner and a low-level controller. The path planner takes from the task server the robot’s current pose, the target point, and the scene map. Using the map, the planner computes a geometric path to the target and passes it to the low-level controller for execution. Path construction uses the Theta* algorithm~\cite{daniel2010theta}, which plans over the occupancy grid in arbitrary directions. The resulting path is a polyline whose segments stay at least a distance $R$ from obstacles marked on the map (the parameter $R$ is chosen based on the robot’s footprint).

The low-level controller~\cite{alhaddad2022adaptive}, implemented in C++ within the ROS framework, manages differential-drive navigation by following a global Theta* path and a local path, both transformed into the robot's coordinate frame, using odometry to compute positional errors at 10 Hz.The path is stored as a sequence of up to 100 waypoints, and the controller tracks the current target waypoint with index \(k_v\). When the robot comes sufficiently close to this waypoint, the index is advanced to the next one. Closeness is defined by the Euclidean distance between the robot’s current odometry position \((x_{odom}, y_{odom})\) and the waypoint coordinates \((x_{qp[k_v]}, y_{qp[k_v]})\):
\begin{equation}
    d_e = \sqrt{(x_{qp[k_v]} - x_{odom})^2 + (y_{qp[k_v]} - y_{odom})^2}.
\end{equation}
If \(d_e < 0.35\) m, the waypoint is considered reached and the controller switches to the next waypoint along the path.
To correct the robot's heading toward the active waypoint, the angular error is defined as:
\begin{equation}
    \theta_e = \arctan2\big(\sin(\alpha - \theta_{odom}), \cos(\alpha - \theta_{odom})\big),
\end{equation}
where
$\alpha = \arctan2(y_{qp[k_v]} - y_{odom}, \, x_{qp[k_v]} - x_{odom})$.  
To ensure continuity, the angular error is wrapped to the interval \([-\pi, \pi]\).

The controller applies a rule-based strategy depending on \(|\theta_e|\). For \(|\theta_e| < 0.1\) rad, the linear velocity control \(v_{cmd}\) increases smoothly toward the maximum \(\max_v = 0.18\) m/s with acceleration \(\,acc_v = 0.04\) m/s\(^2\), while the angular velocity control is defined as \(w_{cmd} = \theta_e\). 

For larger angular errors, the linear velocity limit is reduced and \(w_{cmd}\) is scaled according to the error interval:
\begin{equation}
    w_{cmd} = 
\begin{cases}
0.9 \, \theta_e, & 0.1 \leq |\theta_e| < 0.38, \\
0.85 \, \theta_e, & 0.38 \leq |\theta_e| < 0.7, \\
0.8 \, \theta_e, & 0.7 \leq |\theta_e| < 1.0, \\
0.75 \, \theta_e, & 1.0 \leq |\theta_e| < 1.25.
\end{cases}
\end{equation}

When \(|\theta_e| \geq 1.25\) rad, the robot stops forward motion and rotates in place, with \(|w_{cmd}|\) clamped to the maximum angular velocity \(\max_w = 0.7\) rad/s.

Near-goal deceleration activates in the final segment when \(d_e < 1.5\) m and \(v_{cmd} > 0.3\) m/s, by reducing the local \(\max_v\) at each step. Arrival is flagged when the stop distance satisfies \(d_s \leq 0.09\) m, after which the controller enters final theta correction using the next equation
\begin{equation}
    \theta_e = \arctan2\big(\sin(\theta_{fin} - \theta_{odom}), \cos(\theta_{fin} - \theta_{odom})\big),
\end{equation}
where $\theta_{fin}$ is a desired heading angle at the final moment, and 
commanding fixed-speed rotation \(w_{cmd} = \pm 0.15\) rad/s until \(|\theta_e| < 0.1\).  

\subsection{Manipulation Module}
Our manipulation controller uses Riemannian Motion Policies (RMPs)~\cite{adityacheng2019rmpflowcomputationalgraphautomatic} to compute joint-space accelerations by combining multiple task-space RMPs (attractors, repulsors, joint-limit barriers, and damping). Each RMP is defined on a task map $\Phi_i$ (either the end-effector pose or the joint configuration) with an associated Riemannian metric $M_i$ (defining task-space inertia) and a desired force $F_i$ (encoding the policy behavior). For example, the end-effector position is driven by an attractor towards a target pose, while repulsors provide collision avoidance and additional RMPs in joint space implement joint limit avoidance and velocity damping. End-effector orientation control is handled via quaternion-based feedback to avoid singularities. RMPflow pulls back each task’s $(F_i, M_i)$ from task space to the 8-DoF joint space (parameterized by joint configuration vector $\mathbf{q}$) via the Jacobian $J_i = \nabla_{\mathbf{q}} \Phi_i(\mathbf{q})$, and then combines them using the canonical solver to produce a commanded joint acceleration $\ddot{\mathbf{q}}$. The resulting acceleration is integrated and forwarded as joint velocity and position commands to Isaac Sim’s articulation controller for execution. The ALOHA manipulator has 8 actuated joints (excluding the gripper); the parallel gripper mechanism adds two extra degrees of freedom, which we command with normalized position.

\begin{figure}[t]
    \centering
    \includegraphics[width=\linewidth]{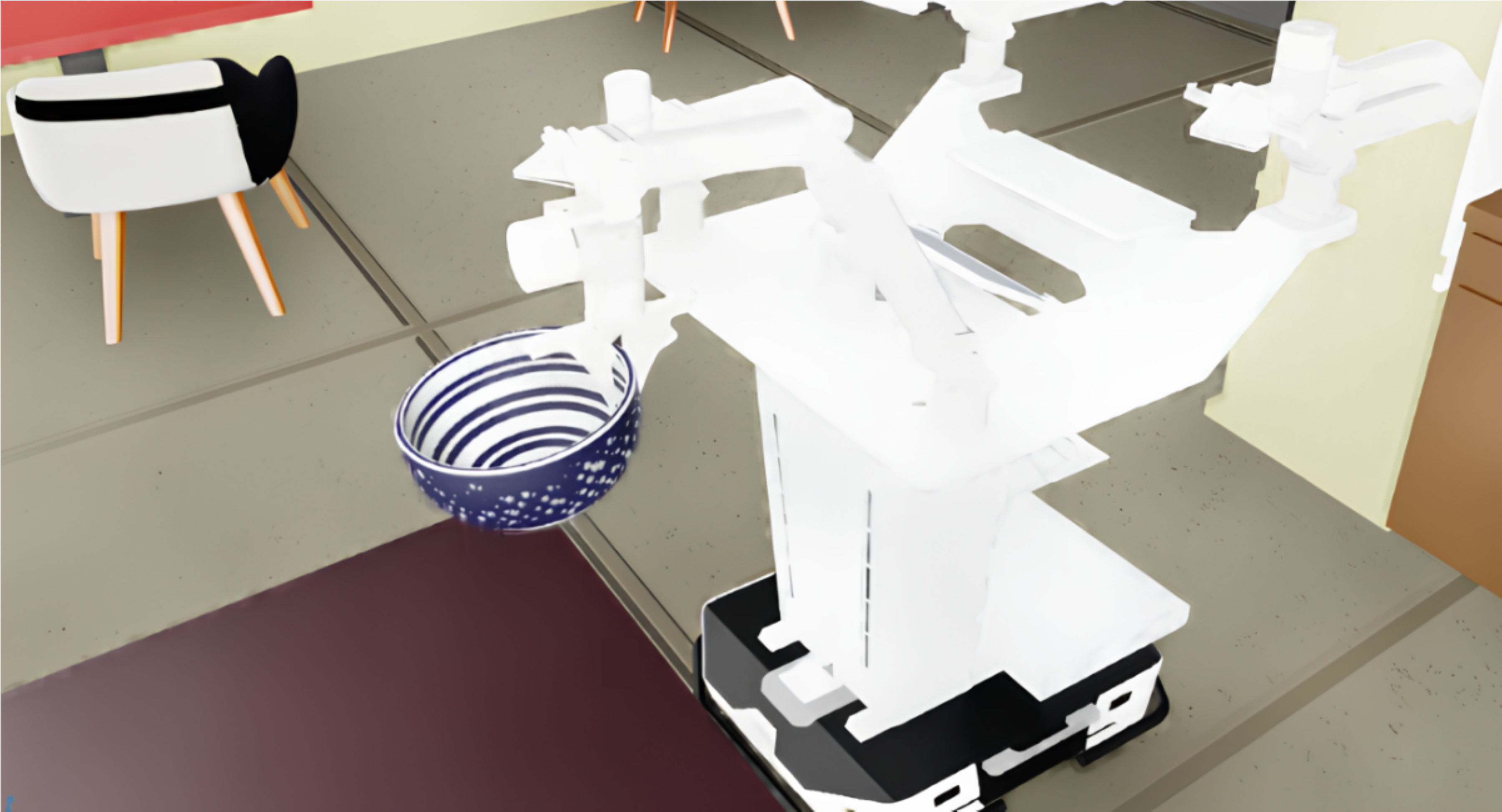}
    \caption{The ALOHA mobile manipulator in simulation executing a ``pick up'' task using the defined RMPflow-based controller. The 10-phase FSM planner guides the end-effector through a sequence of smooth, cosine-blended motions to grasp and lift the target object (a bowl).}
    \label{fig:aloha_pick}
    \vspace{-15pt}
\end{figure}

The manipulation controller operates a 10-phase finite state machine (FSM) to execute a full pick-and-place cycle. Each phase corresponds to a sub-action (e.g., approach object, descend, grasp, lift, transport, release), defined by an end-effector target pose given in the local frame of the relevant scene object. To ensure smooth transitions between phases, we use cosine-blended motion in Cartesian space, a standard interpolation technique \cite{adityacosineinterpolation}. Specifically, we employ a raised-cosine blend $\alpha(\tau) = \frac{1 - \cos(\pi \tau)}{2}$ (with $\tau \in [0,1]$ as normalized time) to interpolate the end-effector’s position (and height) between the current and the next phase targets. This produces smooth, continuous-velocity trajectories with gradual acceleration and deceleration at each phase transition. The orientation reference is similarly interpolated so that the end-effector’s approach and departure are well-aligned with the object. Figure~\ref{fig:aloha_pick} illustrates the ALOHA robot in simulation performing a ``pick up'' task under this controller, following the smooth planned motion.

\subsection{Logging Module}
Proposed framework supports recording data in rosbag and hdf5 formats. Logging to rosbag provides compatibility with standard interfaces of robotic platforms and enables working with simulator data almost as if it were collected on a real robot. The hdf5 format offers advantages such as simplicity of reading/writing, speed, and memory efficiency, but it is not compatible with standard robotics interfaces. To obtain a rosbag in the required format, we use Isaac Sim’s built-in action graphs to publish topics in standard formats, and then ROS republishers to record the data in the desired layout. The pipeline uses five different action graphs to publish various ROS topics: camera, IMU, LiDAR, TF, and odometry. To bridge Isaac Sim with ROS, we use the ROS Bridge extension, which creates virtual topics and forwards data from Isaac Sim in ROS format.

\subsection{Monitoring Module}
To enable real-time validation and collection of high-quality trajectories, data collection framework includes a monitoring and debugging module. While executing a scenario specified by a text plan, it is necessary to monitor the success of basic actions in the simulator: move, pick, place, and to abort an episode if an action fails. For this, the framework contains a set of functions that, at every simulation step, verify the correctness of the actions being executed. Execution time for each basic action is limited by thresholds set in the configuration file. If the time to complete an action exceeds the limit, collection of that trajectory is stopped and the trajectory is marked as unsuccessful. This saves substantial time in large environments (instead of waiting for a robot going the wrong way to drop an object or collide, we abort the run early using prior knowledge of typical action durations). At each step of the overall plan execution, we perform a check for key objects falling to the floor during manipulation. During the final steps of the pick action, we check the distance from a key point on the robot gripper to the target object; it must not exceed a threshold set in the config. After completing place, we check the distance from the object to the goal point. Similarly, for move, we verify that the robot lies within a circle around the goal point; the radius is specified in the task config. 

During scenario execution there may be drops in logging rate, which would produce rosbag trajectories with too low a frequency and negatively affect model training later on. To prevent this, at each step we check the logging rate of every ROS topic. If the rate falls below the configured threshold, the trajectory is marked as unsuccessful and the recording stops. Incorrect collision meshes for key objects can silently cause grasp failures, especially when multiple models are involved. Even with seemingly correct meshes, simulation may fall back to simplified versions after launch, rendering mismatched visuals. Such issues are hard to detect via logs or success rates. To prevent them, we run a preflight collision mesh check before scenario execution. If errors are found, trajectory collection is halted and a warning is issued.

\subsection{Customization and Randomization}
Compared to available alternatives, our framework makes it easy to add new scenes and objects. To use a new scene, it suffices to specify in the config the path to the scene’s USD file and set its scale. To add a new object to a scene, the same parameters are provided. Our framework supports randomization of scene textures, object textures, and poses of both the robot and the scene objects. In each iteration of the trajectory collection, the positions of the key objects for manipulation are randomized relative to a specified point within a circle defined in the config, as are the robot’s positions: its initial zone and a random position along the segment that defines that zone. We also randomize background objects (their models, textures, and positions), which do not directly interact with the robot, as well as scene textures (e.g., floor, walls).

\subsection{Working with Sensors}
To collect multimodal data, proposed framework supports recording RGB-D images from cameras and point clouds from LiDARs. The number of cameras and LiDARs you can add is limited only by the performance of the machine collecting the trajectories. To add a new sensor (e.g., a camera), you need to add it to the robot or scene model with the required parameters, then append the camera primitive path to the cameras list in main config.

\subsection{Creating New Tasks}
Our framework enables creating new tasks from the basic actions: move, pick, place. To define a task, you need a config that includes:
\begin{enumerate}
    \item \textbf{A specified sequence of basic actions.} For example, the task ``Remove the red cup from the table onto the shelf and place the blue cup instead'' can be represented as: move to point 1, pick object 1, move to point 2, place object 1, pick object 2, move to point 1, place object 2.
    \item \textbf{Key points for the basic actions.} For instance, for \textit{move} this is the approach point and the minimum reachability radius; for \textit{pick} this is the pose of the object to grasp and XYZ offsets relative to the nominal pose, taking into account the object’s shape.
    \item \textbf{A set of background and key objects.} For background objects, it suffices to specify a path to a folder with USD files; for key objects, specify the USD model path for each object and the XYZ offsets relative to the nominal pose (accounting for the object’s shape for grasping).
\end{enumerate}

\subsection{Language Instructions}
To evaluate LLM and VLM baselines for task planning, we included 563 natural language instructions in the Kitchen-R benchmark. These instructions pertain to mobile manipulation tasks in a kitchen environment, where the robot can perform basic actions: pick, place, and move to.
The instructions were generated from 58 templates, with the corresponding plans ranging from 4 to 8 steps in length. For each scene in the benchmark, a description of the objects within it is provided. The instruction generation process first selected a scene, then a template and plan, and finally substituted the objects available in that scene into the template. This ensures that every generated instruction is executable within its corresponding scene.

To facilitate the use of VLMs as task planners, a top-down image of each scene is provided. In these images, colored areas mark locations where objects can be placed or from which they can be taken. It is important to note that during execution in the simulator, this color information is not available to the robot's control policy.
An example of an instruction, its corresponding plan, and the top-down scene view are shown in Figure~\ref{fig:dataexample}.

\section{BASELINES}
We also provide baseline methods for our Kitchen-R benchmark: a task planning strategy based on a VLM and a low-level control policy for mobile manipulation based on the Diffusion Policy~\cite{chi2023diffusionpolicy}.
\subsection{VLM-planning Baseline}
\begin{figure}
    \centering
    \includegraphics[width=\linewidth]{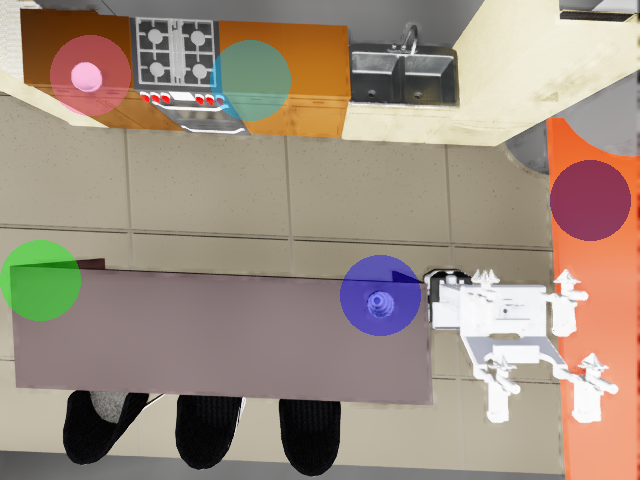}
    \caption{Top-down view of the planning environment (instruction and expected output are ''\texttt{INPUT: Move the bowl to the green zone. TARGET: move to the blue zone, pick the bowl, move to the green zone, place the bowl}'').}
    \label{fig:dataexample}
\end{figure}

As a baseline algorithm for planning, we went with the following pipeline. The open source VLM OmniFusion~\cite{omnifusion} operated on initial top-down views of the kitchen scene along with visual cues and instructions (Figure~\ref{fig:dataexample}), expressed in natural language, and produced sequences of tasks that were interpreted as atomic instructions by the mobile manipulation baseline. To ensure adequate performance, we provided the model with in-context examples of correct plans, as well as a set of valid low-level instructions. Since the model has been pre-trained on datasets with only a single-image dialogues, our in-context examples include textual description (by OmniFusion) instead of actual training dataset images. Also, constrained text generation \cite{constrained_generation} was used to further reduce risk of generating uninterpretable instructions. Ablations of different components of the planning algorithm are presented in Table~\ref{tab:planning_baseline_comparison} where performance is measured by the Exact Match metric between the generated plan and ground truth.

\begin{table}[t]
    \centering
    \caption{Effect of different components of the VLM-planning baseline on the Exact Match metric.}
    \begin{tabular}{l|c}
            \toprule
            Method & Exact Match $\uparrow$ \\
            \midrule
            Vanilla OmniFusion & 0.000 \\
            + Valid instructions added to context & 0.000 \\
             + In-context examples of plans & 0.612 \\
             + Constrained generation & \textbf{0.632} \\
            \bottomrule
        \end{tabular}
    
    \label{tab:planning_baseline_comparison}
\end{table}

\subsection{Mobile Manipulation Baseline}
The architecture of the mobile manipulation policy is based on the Diffusion Policy framework~\cite{chi2023diffusionpolicy}. The conditions for the diffusion model include feature extraction from two camera images with a history window size of 2, and the robot state, represented by ten coordinates relative to the previous robot state coordinate system (three for translation of the gripper tips, six for 6 degrees of freedom, and one for gripper gap), with a history window size of 2. To effectively capture the features of the surrounding environment, we utilized spatial features from the backbone with a compression layer as proposed by~\cite{shang2024theia}. 

The robot state is encoded using a linear layer and concatenated with image features to form a condition sequence. Vision backbones do not share weights for different camera views and use an updating weight of 10\% of the learning rate during training, which resulted in enhanced policy quality. We modified the original FiLM~\cite{perez2018film} conditioning by replacing it with cross-attention blocks. The condition is structured as a sequence of tokens and employed in the cross-attention blocks of the U-Net~\cite{ronneberger2015u} model during the diffusion process to denoise the robot action sequence over 16 future timesteps.

\section{BASELINE VALIDATION AND RESULTS}
To evaluate policies and plans, we compute the success rate and execution time of a policy on a set of tasks, where each task is defined by an environment configuration and a high-level action plan. Environment allows to control the Mobile ALOHA robot by providing linear and angular velocities of the robot's base, 3D coordinates and quaternion for the desired gripper position and orientation, and a command to open or close the gripper.
The validation process consists of a set of independent episodes, performance metrics are computed for each episode and then averaged. The validation process proceeds in four steps.

\textbf{Step 1: Define Distribution Parameters.} These parameters shape the generation of environment configurations, which serve as the initial states for episodes. The parameters include:
\begin{itemize}
    \item \textbf{a set of line segments}, over whose union a uniform distribution is defined for generating the robot's initial position in an episode;
    \item \textbf{a set of manipulation objects:} \{apple, bowl, cup, fork, plate\}, from which one object is selected for the entire episode and instantiated as a rigid body;
    \item \textbf{a set of five points}, in the vicinity of which the initial and target final positions of the manipulation object are chosen. Two points -- a start and a goal -- are selected per episode;
    \item \textbf{the radius of the vicinity} around the start and goal points from which the object's initial and target positions are sampled.
\end{itemize}

\textbf{Step 2: Generate Environment Configuration.} A single configuration is generated for each episode. If a set of natural language commands is provided to assess their feasibility, the object type and its initial and target positions are fixed during generation as they are determined by the given command. If no command is specified (e.g., during general policy validation), the object type and its positions are sampled from their respective distributions. A corresponding set of commands is then generated based on the environment configuration and remains fixed for the episode. For debugging or demonstration purposes, all distribution parameters can be set manually, or environment configuration attributes can be assigned desired values directly. To compare the performance of a set of policies on the same set of episodes, it is possible to pre-generate a batch of configurations and save them. This allows for controlling the variation of any parameter across the generated configurations, for instance, to achieve a uniform distribution of object types over a small set of configurations.

\textbf{Step 3: Create Environment.} Based on the generated configuration, an environment with a standard OpenAI Gym interface is created.

\textbf{Step 4: Execute Evaluation.} The performance evaluation of the given policy begins. First, the environment is reset to the initial state defined by the provided configuration. Then, in a loop, the policy receives an observation from the environment and uses it to generate an action. This action is passed to the environment and influences the outcome of the next simulation step. The observation includes a decomposed task in natural language, as well as visual information from two cameras mounted on the robot (one on the gripper and a central camera on the base). The simulation for the current task terminates upon successful completion by the policy or when a predefined time limit for the task is reached. After that, the validation continues for the next configuration, and the results are written to a file. After all episodes are completed, the performance metrics for each policy are averaged and reported as the final result of the validation process.

In addition to time limits, the validation process can be configured to record video of an episode, teleport the robot to the start of the next task if the current one is failed, and specify the number of actions from a model-generated plan to execute in the environment. The result of a single validation episode includes success flags for all sub-tasks and their execution durations as measured in simulation time. The policy's inference time does not affect the simulation time. During validation, detailed logs are recorded for each episode, allowing for process monitoring without requiring visual rendering.

The execution time for a single episode on an NVIDIA GeForce RTX 3060 Ti GPU is approximately 1 minute when using the oracle. This can increase up to 50 minutes for policies with high inference times.

\section{CONCLUSION}
We introduced Kitchen-R, a benchmark that bridges the gap between high-level instruction planning and low-level control evaluation in Embodied AI. This digital twin kitchen, with its diverse instructions and realistic simulation, enables integrated and modular assessment of full robotic systems. Kitchen-R was used to collect approximately 2,700 mobile manipulation trajectories and over 500 diverse planning language instructions for the Embodied AI track of the 2024 AIJ Contest. By providing a unified testbed and baselines, Kitchen-R facilitates the development of robust agents that can effectively execute complex language commands in realistic environments.

\bibliographystyle{IEEEtran}
\bibliography{IEEEexample}
\end{document}